\documentclass[12pt]{iopart}

\usepackage{array}
\usepackage{tabularx}
\usepackage{graphicx} 
\usepackage{cite}
\usepackage{siunitx}
\usepackage[colorlinks=true,linkcolor=blue]{hyperref}
\expandafter\let\csname equation*\endcsname\relax
\expandafter\let\csname endequation*\endcsname\relax
\usepackage{amsmath}
\usepackage{threeparttable}

\begin{document}

\title[STEMNIST]{STEMNIST: Spiking Tactile Extended MNIST Neuromorphic Dataset}
\author{Anubhab Tripathi$^{1}$, Li Gaishan$^{1}$, Zhengnan Fu$^{1}$, Chiara Bartolozzi$^{2}$, Bert E. Shi$^{3}$ and Arindam Basu$^{1,4,*}$}
\address{$^{1}$ Department of Electrical Engineering, City University of Hong Kong, 83 Tat Chee Avenue, Kowloon, Hong Kong}
\address{$^{2}$ Event-Driven Perception for Robotics, Istituto Italiano di Tecnologia, Via Morego 30, 16163 Genoa, Italy}
\address{$^{3}$ Department of Electronic and Computer Engineering, Hong Kong University of Science and Technology, Clear Water Bay, Kowloon, Hong Kong}
\address{$^{4}$State Key Laboratory of Terahertz and Millimeter Waves, City University of Hong Kong, 83 Tat Chee Avenue, Kowloon, Hong Kong}
\address{$^{*}$ Author to whom any correspondence should be addressed.}
\ead{arinbasu@cityu.edu.hk}

\date{October 2025}

\begin{abstract}
Tactile sensing is essential for robotic manipulation, prosthetics and assistive technologies, yet neuromorphic tactile datasets remain limited compared to their visual counterparts. We introduce STEMNIST, a large-scale neuromorphic tactile dataset extending ST-MNIST from 10 digits to 35 alphanumeric classes (uppercase letters A–Z and digits 1–9), providing a challenging benchmark for event-based haptic recognition. The dataset comprises 7,700 samples collected from 34 participants using a custom 16$\times$16 tactile sensor array operating at 120 Hz, encoded as 1,005,592 spike events through adaptive temporal differentiation. Following EMNIST's visual character recognition protocol, STEMNIST addresses the critical gap between simplified digit classification and real-world tactile interaction scenarios requiring alphanumeric discrimination. Baseline experiments using conventional CNNs (90.91\% test accuracy) and spiking neural networks (89.16\%) establish performance benchmarks. The dataset's event-based format, unrestricted spatial variability and rich temporal structure makes it suitable for testing neuromorphic hardware and bio-inspired learning algorithms. STEMNIST enables reproducible evaluation of tactile recognition systems and provides a foundation for advancing energy-efficient neuromorphic perception in robotics, biomedical engineering and human-machine interfaces. The dataset, documentation and codes are publicly available to accelerate research in neuromorphic tactile computing.
\end{abstract}

\section{Introduction}
Haptic Perception\cite{Lederman2009HapticTutorial} and Sensing have long intrigued scientists, as it is one of the most important modalities that helps us understand the surroundings when other avenues of senses such as vision and audition are blocked or inaccessible. One of its main components is ``Touch"\cite{gibson1962observations} -- one of the most vital human senses, enabling our natural interaction with the environment through direct contact and deformation sensing. Tactile sensing is of great importance in making it possible to develop artificially intelligent machines that are not only able to perform dexterous object manipulation safely but also engage in Human Computer Interaction\cite{jin2023progress} proficiently by providing real-time information about contact dynamics and material properties. Resistive Tactile Arrays\cite{roberts2021soft,wangwei_kilo} stand out among the current sensor technologies in robotics research due to their scalability and affordability, making large scale deployments possible. Pressure applied to the arrays of sensor elements causes a large change in resistance that can be scanned, digitized and processed further for downstream tasks.

Pioneering work such as \cite{wangwei_kilo} demonstrated the benefits of neuromorphic approaches in capturing tactile data with high temporal resolution, an aspect that neuroscientists believe to be responsible for fast sensory responses in biology\cite{tactile_temporal_neurosci}. However, there is a lack of large-scale datasets in event-driven neuromorphic domains that can help quantitatively assess the difference in information carried by high temporal-precision tactile data. This is even more important now given the recent surge of artificial tactile systems in robotics and biomedical engineering. The seminal MNIST\cite{lecun2002gradient} dataset enabled transformative advances in machine learning for handwritten digit recognition and its extension EMNIST\cite{cohen2017emnist} similarly expanded benchmarking to the entire English alphabet through standardized data splits and tasks. However, neuromorphic tactile analogues have been limited. ST-MNIST\cite{see2020st} introduced the first publicly available neuromorphic tactile dataset, capturing asynchronous event-based signatures of digit handwriting on a 10$\times$10 array and reporting both Convolutional Neural Network (CNN) and Spiking Neural Network (SNN) baselines, establishing the viability of spatio-temporal tactile patterns for efficient event-driven classification. Despite this, ST-MNIST is limited to just digits (0 to 9) and a lower spatial resolution compared to modern tactile skins, restricting its utility for learning the greater variety of spatio-temporal forms found in alphabetic symbols. In contrast, the STAG\cite{sundaram2019learning} dataset delivered large-scale tactile videos from a human hand equipped with a 548-sensor glove interacting with diverse objects, achieving rich spatial coverage for full-hand grasp analysis but operating at a substantially lower frame rate ($\sim$7.3 Hz), which blurs the fine event timing information necessary for neuromorphic encoding and hinders timing-sensitive spiking inference. More recently, the Braille letter reading dataset\cite{muller2022braille} recorded spatio-temporal patterns of 27 Braille classes (letters A to Z and space) using the iCub robot's fingertip with 12 capacitive tactile sensors sampled at 40 Hz, demonstrating neuromorphic hardware inference on Intel Loihi with high energy efficiency partially filling this requirement. Yet its minimal spatial resolution and limited temporal sampling remain insufficient for capturing the full complexity of more complex tactile patterns. Table \ref{tab:tactile_datasets} compares the details of these datasets.

\begin{table}[htbp]
\centering
\caption{Comparison of Event-Based Tactile Datasets}
\label{tab:tactile_datasets}
\begin{tabular}{@{}l>{\raggedright\arraybackslash}p{2.3cm}>{\raggedright\arraybackslash}p{2.5cm}>{\raggedright\arraybackslash}p{1.8cm}>{\raggedright\arraybackslash}p{4cm}@{}}
\hline
\textbf{Dataset} & \textbf{Modality} & \textbf{Classes} & \textbf{Scale} & \textbf{Notable Characteristics} \\ 
\hline
ST-MNIST \cite{see2020st} & Event-based tactile array & 10 digits & 6,953 samples & 10$\times$10 taxels, FA-like responses, 2-second recording, ANN/SNN baselines \\ 
\hline
STAG \cite{sundaram2019learning} & Synchronous tactile glove & 26 objects & 135,187 frames & 548 sensors, $\sim$7.3 Hz, object ID and weight estimation \\ 
\hline
Braille \cite{muller2022braille} & Capacitive tactile sensors & 27 (A-Z and space) & 5,400 samples & 12 taxels, 40 Hz, iCub fingertip, Sigma-delta encoding,  Loihi \\
\hline
\parbox[t]{2.3cm}{STEMNIST \\ (this work)} & Event-based tactile array & 35 characters (26 letters, A-Z and 9 digits, 1-9) & 7,700 samples 1,848,000 frames & 16$\times$16 taxels, 120 Hz, 2-second recording, Differential-frame preprocessing, NN baselines \\ 
\hline
\end{tabular}
\end{table}

To address these complementary gaps, we present STEMNIST, a neuromorphic tactile dataset that extends tactile event recording to include handwritten uppercase English Alphabet letters (A to Z) along with numerical digits (1 to 9), captured on a 16$\times$16 Tactile Sensor Array (TSA) with high temporal resolution (120 Hz) and accurate event-like preprocessing. To guarantee reproducibility and equitable benchmarking for both ANN and SNN approaches, STEMNIST leverages neuromorphic-friendly differential encoding and adheres to standardized, EMNIST-inspired task partitions. This resource is designed to catalyze progress in tactile perception for the robotics, biomedical engineering and neuromorphic computing communities by providing the necessary breadth of classes and spatio-temporal fidelity for benchmarking advanced learning algorithms, enabling research in tactile interfaces, assistive devices and beyond.

The rest of the paper is organized as follows: Section \ref{sec:related} provides a deeper exploration of prior work. In Section \ref{sec:hardware}, we describe the tactile hardware system and elaborate on the data collection process in Section \ref{sec:collection}. We offer a comprehensive discussion of the dataset in Section \ref{sec:dataset}. Classification results are reported in Section \ref{sec:classification}. Section \ref{sec:future} delineates potential directions for future research, followed by the conclusion in Section \ref{sec:conclusion}.

\section{Related Work}\label{sec:related}
The MNIST dataset established a baseline standard for handwritten digit classification and gave rise to numerous Deep Learning methods and acted as a test bed for neural network architectures. EMNIST then built on this foundation systematically by adding letters sourcing from NIST Special Database 19\cite{grother1995nist} to incrementally increase the class vocabulary and standardizing task splits for equitable evaluation and comparison studies. However, these frame-based datasets fundamentally discard temporal information by representing handwriting as static spatial patterns, limiting their utility for evaluating neuromorphic computing systems that exploit spike-timing-dependent processing. To bridge this gap, neuromorphic adaptations of MNIST emerged by converting static images into event streams using bio-inspired vision sensors. MNIST-DVS\cite{serrano2015poker} employed a Dynamic Vision Sensor (DVS) to record pixel-level brightness changes as images were displayed on an LCD monitor with controlled motion, generating asynchronous address-event representation (AER) spike trains. Similarly, N-MNIST\cite{orchard2015converting} utilized an Asynchronous Time-based Image Sensor (ATIS)\cite{posch2008asynchronous} to encode MNIST digits through synthetic saccadic eye movements, enabling evaluation of spiking neural networks on temporal event data.

\begin{figure}[htb]
\centering
\includegraphics[width=\textwidth]{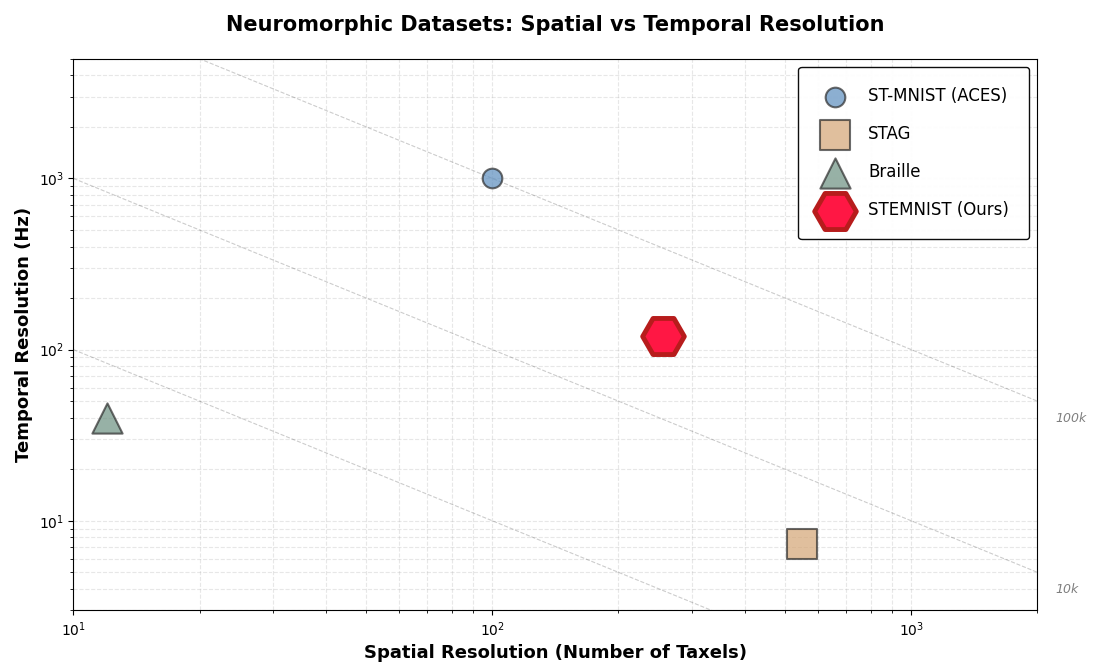}
\caption{Comparison of Neuromorphic Spatio-Temporal Datasets.}
\label{fig:neuro_spatiotemporal}
\end{figure}

However, the temporal richness of these synthetically generated neuromorphic datasets has been called into question. Iyer et al.\cite{iyer2021neuromorphic} demonstrated that frame based artificial neural networks trained on time-collapsed versions of N-MNIST and N-Caltech101 achieved 99.23\% and 78.01\% accuracy respectively matching or exceeding state-of-the-art spiking network performance on the same datasets. By comparing spike-timing-dependent plasticity algorithms (RD-STDP for spatial-only processing and STDP-tempotron for spatiotemporal processing), they showed that N-MNIST could be adequately classified using spatial features alone, while truly dynamic datasets like DvsGesture\cite{amir2017low} required temporal algorithms. This finding reveals a fundamental limitation: converting static images to event streams through artificial camera movements does not guarantee the presence of discriminative temporal information that would justify the computational overhead of spike-timing-dependent processing in SNNs. Such synthetic datasets may therefore underestimate the potential advantages of neuromorphic architectures, which are designed to exploit millisecond-precise temporal dynamics rather than spatial patterns distributed across artificially elongated time windows.

These observations motivate the development of neuromorphic benchmarks derived from inherently dynamic sensory modalities where temporal structure emerges naturally from physical interaction rather than synthetic methods. Tactile sensing during handwriting represents an ideal candidate: pressure changes unfold continuously as a finger traces character strokes across a sensor array, encoding genuine temporal dependencies related to writing speed, stroke order, and contact dynamics features absent in static visual images. Unlike vision-based datasets where temporal information is retroactively imposed on spatially-complete images, tactile handwriting intrinsically produces sparse, asynchronous events distributed across space and time, aligning naturally with event-driven neuromorphic processing paradigms. Existing tactile datasets although have not fully realized this potential. High-density tactile arrays for robotic manipulation\cite{kong2025super} and sensor architectures with sophisticated analog front-ends have advanced spatial resolution and material science, but publicly available benchmarks combining event-based encoding, high spatiotemporal resolution, and substantial class complexity for character-level recognition remain scarce. ST-MNIST\cite{see2020st} introduced the first neuromorphic tactile character dataset with 10 handwritten digits on a 10$\times$10 array, reporting both CNN and SNN baselines to establish the viability of event-driven tactile classification. Yet, its restriction to digits and modest spatial resolution limit its utility for evaluating neuromorphic algorithms on the diverse spatiotemporal patterns required for full alphanumeric recognition. Although the Braille letter reading dataset \cite{muller2022braille} extended the class count to 27 Braille characters (A-Z and space), the minimal spatial resolution (12 taxels) and limited temporal sampling (40 Hz) constrain its applicability to more complex tactile interaction scenarios requiring finer-grained spatio-temporal discrimination. Neuromorphic computing offers transformative advantages for scalable, low-power embedded sensing through asynchronous, sparse event-driven coding of activity-dependent spatiotemporal streams. Realizing this potential requires datasets preserving natural tactile dynamics, substantial class complexity, and standardized evaluation. Existing neuromorphic vision datasets lack this sensorimotor richness, relying on synthetic temporal structure. STEMNIST addresses this by capturing human handwriting's intrinsic spatiotemporal complexity via direct tactile measurement at high spatial (256 taxels) and temporal (120 Hz) resolution, encoding 35 alphanumeric classes (A-Z, 1-9) as asynchronous spike events via adaptive temporal differentiation extending ST-MNIST's challenge while enabling tests of temporal coding advantages in organically dynamic domains.

\section{Materials and Methods}
\subsection{Neuromorphic Tactile Sensing Hardware System}\label{sec:hardware}

\begin{figure}[htb]
\centering
\includegraphics[width=\textwidth]{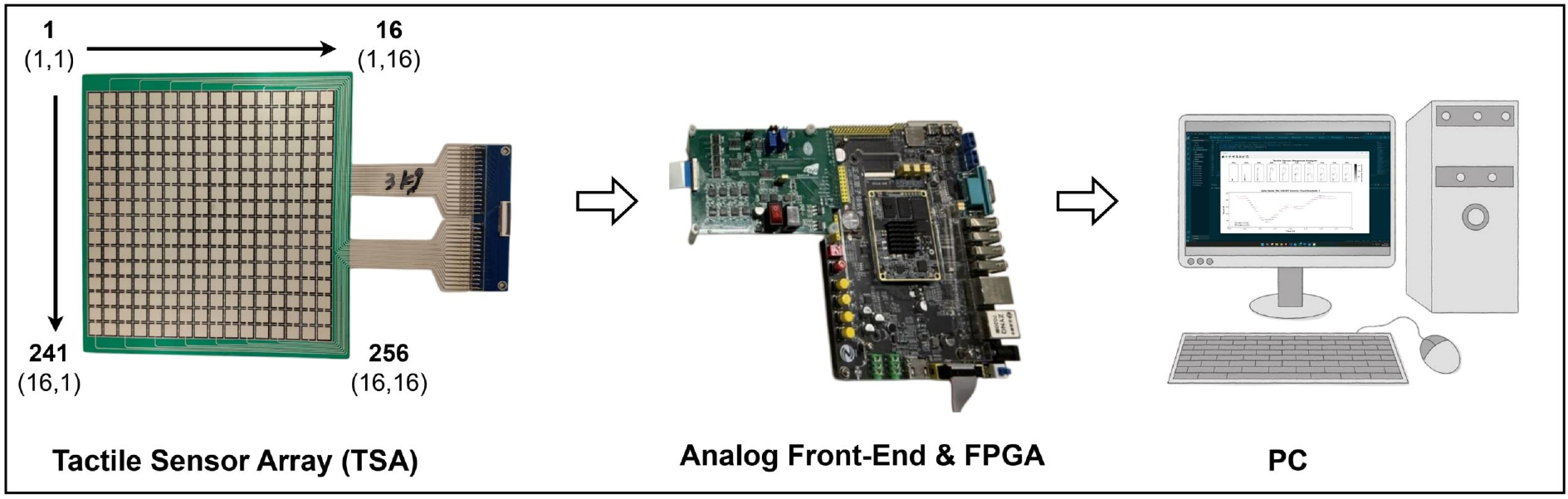}
\caption{Tactile Sensing Hardware System used to create the dataset comprising a $16\times 16$ tactile sensor array (TSA), readout electronics and FPGA for data transfer to a PC.}
\label{fig:hardware}
\end{figure}

Figure \ref{fig:hardware} illustrates the distributed flexible pressure sensor array, which is fabricated by precisely printing nanomaterials-based force-sensitive layers and silver interconnects on a thin-film substrate. The sensor consists of 256 units, each with dimensions of 7.5 mm $\times$ 7.5 mm and is configured in a 16$\times$16 array on an area of 150 mm $\times$ 150 mm. When the applied pressure is increased, the resistance decreases as a result of piezo-resistive behavior, which is confirmed by calibration using a robotic arm. In fact, resistance decreases monotonically from several G\si{\ohm} to 3 k\si{\ohm} during the tested range, showing the inverse relationship to the applied pressure. This well-defined feature guarantees an appropriate dynamic range and readout resolution, providing critical guidance on the design of the front-end amplifier.

\begin{figure}[htb]
\centering
\includegraphics[width=\textwidth]{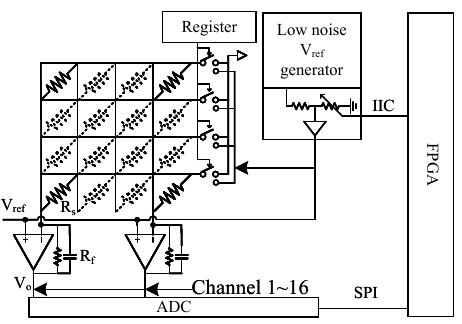}
\caption{Schematic of the Readout Hardware.}
\label{fig:schematic}
\end{figure}

Figure \ref{fig:schematic} encapsulates a multi-channel data acquisition system built around a Field Programmable Gate Array (FPGA) and the 16$\times$16 resistive Tactile Sensor Array. Each sensor element is integrated into a parallel negative feedback circuit. The system employs switch chips (ADG734) controlled by a register (74HC16D) to connect array rows either to a precision reference voltage $V_{ref}$ or to ground (GND). A feedback amplifier configuration using OPA2991 operational amplifiers is activated by connecting a row to GND. This switchable architecture provides for several monitoring modes and enhances the data compression capability of the system. The output voltage is defined by $V_o = V_{ref}(1 + R_f / R_s)$, where the feedback resistor $R_f$ = 100 k\si{\ohm} and the sensor resistance $R_s$ ranges from several k\si{\ohm} up to hundreds of G\si{\ohm}. For Miller compensation, a compensation capacitor $C_f$ is placed across $R_f$ to offer stability against parasitic capacitance effects originating from the high impedance source. 

To fully utilize the dynamic range of the ADC, the signal chain employs eight low-noise, rail-to-rail OPA2991 op-amps. A high-stability REF5025 voltage reference with a temperature coefficient of 3 ppm/$^\circ$C is used to generate the reference $V_{ref}$. The amplified analog signals are digitized by a 16-channel ADS7961 ADC with a programmable input range of 0 to $V_{ref}$ or 0 to $V_{ref}/2$, which can sample at 1 MHz. The digital data is then transmitted to the FPGA for further processing via the Serial Peripheral Interface (SPI) communication protocol.

\subsection{Data Collection}\label{sec:collection}
34 participants participated in the trials, with each one generating 35 distinct character classes composed of handwritten samples of uppercase English letters (A to Z) and numerical digits (1 to 9). The participants were asked to inscribe the characters directly onto the surface of the 16$\times$16 TSA using their index finger. The collection protocol divided these 35 characters into seven consecutive sets of five characters each for systematic and balanced data acquisition. The seven sets were Set 1 (A to E), Set 2 (F to J), Set 3 (K to O), Set 4 (P to T), Set 5 (U to Y), Set 6 (Z, 1 to 4) and Set 7 (5 to 9). Each participant completed all seven sets in sequence and this full cycle was repeated 10 times for 10 participants initially. Afterwards, the number of repetition was reduced to 5 times, yielding 5 samples per character per participant. Adequate breaks were provided after the completion of each cycle to prevent the onset of fatigue upon the participators. This ensured consistency in quality of collected data throughout the recording sessions. 

\begin{figure}[htb]
\centering
\includegraphics[width=\textwidth]{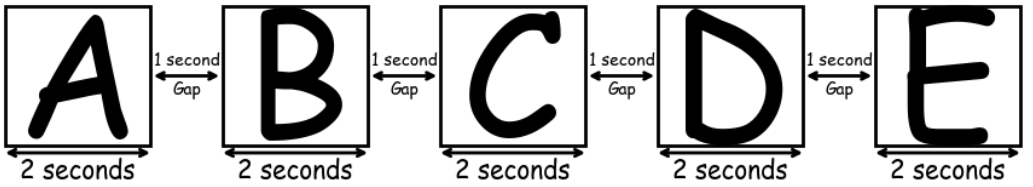}
\caption{Protocol of Data Collection.}
\label{fig:protocol}
\end{figure}

To maintain precise time intervals and provide clear and easy to interpret visual feedback for participants, a Light-Emitting Diode (LED) indicator controlled by the FPGA was programmed to illuminate for exactly 2 seconds for each character, signaling the active data acquisition window during which the TSA was being scanned. Following each 2-second recording period, a 1-second inter-trial interval was provided to signal the participant to prepare for the next character, during which the LED turned off before illuminating again for 2 seconds to record the next character. The protocol for recording 5 characters of Set 1 is illustrated in Figure \ref{fig:protocol}. The LED cues ensured all participants were generally starting their response at the same time and were aware of the exact recording time, thus contributing to consistent timing and reducing variability in participant reaction times due to any timing uncertainty.

The TSA was continuously scanned at a sampling rate of 120 Hz throughout each 2-second recording window, capturing high temporal-resolution pressure distribution data across all 256 taxels (16$\times$16 array). During data acquisition, participants wrote five characters sequentially within each set while the FPGA-controlled data acquisition system streamed raw sensor readings to a host computer via SPI at baud rate of 115200. Raw ADC outputs from the ADS7961 were digitized as 8-bit unsigned integers. Due to the piezo-resistive sensor characteristics, amplifier gain configuration and baseline reference voltage $V_{ref}$  optimized for the tested pressure range, the observed output values during active handwriting exhibited an effective dynamic range predominantly between 75 and 255, with lower baseline values ($\sim$75) corresponding to no applied pressure and higher values (approaching 255) corresponding to strong contact pressure. 

To ensure data integrity, the acquisition pipeline used a specially designed Graphical User Interface (GUI) based on Python that synchronized the LED timing and serial data capture with real-time validation of the frame markers (0x0F hexadecimal delimiter). The five-character recordings within each set were initially stored as a single HDF5 file (1,200 frames total per set) with metadata including capture timestamp, sampling rate and array dimensions. Then, using an automated post-processing Python script that evenly splits the frame sequence to balance temporal windows across characters, each five-character HDF5 was segmented into individual character samples, creating five separate files, each containing the frames for one character. The resulting dataset consists a total of 7,700 samples with 1,848,000 frames in total, distributed evenly across all 35 character classes to ensure equal class representation. Each sample was stored as an individual Hierarchical Data Format version 5 (HDF5) file in a centralized raw character repository with a standardized naming convention \verb|<ParticipantID>_<Character>_<RepetitionNumber>.h5|, facilitating subsequent event-based preprocessing and neuromorphic encoding. Raw pressure data files were stored at full temporal resolution (120 Hz) to allow for flexible threshold-based spike generation and differential frame analysis, as detailed in Section \ref{sec:dataset}.

\section{Spiking Tactile Extended MNIST Dataset}\label{sec:dataset}
\subsection{Spike Generation and Encoding}
Raw pressure frames recorded at 120 Hz were post-processed to generate event-like spike representations suitable for neuromorphic computing. Spike generation employed temporal differentiation between consecutive frames, where pressure increasing above a threshold generated ON spikes (polarity $+1$) and decreasing below the corresponding negative threshold generated OFF spikes (polarity $-1$).

The temporal differential for each sample at frame $f$ was computed as:
\begin{equation}
    \Delta P(f, x, y) = P(f, x, y) - P(f-1, x, y)
\end{equation}
where $P(f, x, y)$ is the pressure value at frame $f$ and taxel position $(x, y)$ of the 16$\times$16 array. To accommodate inter-participant variability in writing pressure, we applied an adaptive per-sample thresholding scheme. For each sample, the threshold $\theta_{\text{sample}}$ was calculated as the 90th percentile of its temporal differential distribution, computed over non-zero absolute values:
\begin{equation}
    \theta_{\text{sample}} = \operatorname{percentile}_{90}\bigl(\{\,|\Delta P(f,x,y)| : |\Delta P(f,x,y)| \neq 0\,\}\bigr)
\end{equation}
ensuring that the upper decile (top 10\%) of observed non-zero absolute pressure change magnitudes were classified as spike events. Zero values (representing taxels with no pressure change) were excluded from the percentile calculation because tactile handwriting is spatially and temporally sparse, typically activating few taxels at any given time. Since the vast majority of temporal differences are zero, computing percentiles over the full distribution would yield thresholds dominated by background inactivity rather than the magnitude distribution of actual tactile events. This exclusion ensures the threshold reflects stroke dynamics occurring along active trajectories. To prevent noise, the computed threshold was constrained to a minimum value of 4:
\begin{equation}
    \theta_{\text{sample}}^{\text{clipped}} =  \max(\theta_{\text{sample}}, 4)
\end{equation}


This lower bound corresponds approximately to the sensor's noise floor below which spurious events from electronic noise dominate genuine tactile signals. No upper bound was applied, allowing the adaptive method to naturally accommodate the full range of writing pressures encountered across participants from light touch to heavy pressure. This percentile-based clipping strategy, analogous to robust normalization methods for event-driven neural networks\cite{rueckauer2017conversion} maintains sufficient spike rates to preserve temporal dynamics for most samples. Conservative outlier-based normalization (e.g., using per-sample maxima without bounds) can lead to insufficient firing rates for the majority of samples, whereas percentile-based bounds ensure consistent event generation across diverse writing styles. Across the entire dataset of 7,700 samples, the adaptive threshold had a mean, $\mu = $ 7.77 and standard deviation, $\sigma = $ 4.44, confirming effective adaptation to diverse writing styles while maintaining a noise-robust lower bound.

Spikes were then generated according to the following criterion:
\begin{equation}
    \text{Spike}(f, x, y) = 
    \begin{cases}
        +1 & \text{if } \Delta P(f, x, y) > \theta_{\text{sample}}^{\text{clipped}} \quad \hspace{2ex}(\text{ON spike}) \\
        -1 & \text{if } \Delta P(f, x, y) < -\theta_{\text{sample}}^{\text{clipped}} \quad (\text{OFF spike}) \\
        0 & \text{otherwise} \hspace{18ex} (\text{NO spike})
    \end{cases}
\end{equation}

Each spike event was encoded as a tuple $(t_s, \text{ID}_{\text{taxel}}, p)$ in accordance with the Address Event Representation (AER) format standard for neuromorphic systems. Here, $t_s$ represents the spike timestamp in seconds, calculated as $t_s = f / f_{\text{sample}}$ where $f$ is the frame index and $f_{\text{sample}} = 120$ Hz is the sampling rate. The taxel identifier $\text{ID}_{\text{taxel}} \in [1, 256]$ is computed as:
\begin{equation}
    \text{ID}_{\text{taxel}} = (X - 1) \times 16 + Y
\end{equation}
for row $X$ and column $Y$ (both 1-indexed), and $p \in \{-1, +1\}$ denotes spike polarity. This sparse, event-driven representation preserves the asynchronous temporal structure of tactile handwriting while enabling efficient storage and processing on neuromorphic hardware.

\subsection{File Format and Organization}\label{sec:format}
The STEMNIST dataset has a hierarchical directory structure to support both spike-based neuromorphic processing and frame-based deep learning methods. The dataset consists of three main components:

\textbf{Raw Pressure Data:} The raw 8-bit pressure data associated with each handwritten character, initially consolidated within a single five-character HDF5 file, were subsequently disaggregated and saved as individual HDF5 files through Python-based post-processing. Each sample comprises a three-dimensional array of pressure readings (240 frames $\times$ 16 $\times$ 16) accompanied by comprehensive meta-data, including participant identifier, character label, repetition index, sampling frequency (120 Hz), and temporal acquisition timestamp. The file naming convention adheres to the scheme detailed in Section \ref{sec:collection}, following the structure \texttt{<ParticipantID>\_<Character>\_<RepetitionNumber>.h5}.

\textbf{Processed Spike Data:} These are event-based data representations, stored in HDF5 files as structured NumPy arrays. Three fields are present in each spike file: timestamp (float32), taxel ID (int32) and polarity (int16). Additionally, each spike file contains attributes that display the total number of spikes, the adaptive threshold value ($\theta_{\text{sample}}^{\text{clipped}}$) and a reference to the original raw file.  Spike files use the naming convention \texttt{<ParticipantID>\_<Character>\_<RepetitionNumber>\_spikes.h5} and are arranged into 35 subdirectories according to their respective character class.

\textbf{Train-Test Split:} Classification experiments use stratified random 80-20 train-test splits that are repeated over five independent trials with 5 random seeds (42, 123, 456, 789, 1011), ensuring balanced class distribution across splits. To evaluate model generalization consistency and robustness, results are presented as mean accuracy $\pm$ standard deviation over the five trials in Section \ref{sec:classification}. The full set of samples post-spike generation is included in the \texttt{ProcessedSpikes} directory, where all spike files are further arranged by character class for reference and reproducibility.

\subsection{Dataset Properties}
The STEMNIST dataset includes 7,700 samples evenly spread across 35 character classes (26 uppercase letters A–Z and 9 digits 1–9), with 220 samples per class. Each sample contains a 2-second recording (240 frames at 120 Hz) from the 16$\times$16 TSA, totaling 61,440 raw pressure measurements per sample. Following adaptive spike generation, the dataset contains 1,005,592 spike events in total, with 498,230 ON spikes and 507,362 OFF spikes, yielding an average of 130.60 spikes (64.7 ON and 65.9 OFF) per sample.

\begin{table}[htbp]
\centering
\small
\begin{threeparttable}
\caption{Descriptive statistics for Letters A to E}
\label{tab:stemnist_stats_1}
\begin{tabular}{@{}lccccc@{}}
\hline
Character & \textbf{A} & \textbf{B} & \textbf{C} & \textbf{D} & \textbf{E} \\
\hline
ON events & 64.76 (17.26) & 78.05 (23.49) & 59.55 (20.28) & 67.85 (22.27) & 69.96 (22.10) \\
OFF events & 65.21 (16.88) & 78.56 (22.91) & 58.54 (19.22) & 68.05 (21.43) & 72.46 (23.94) \\
X mean & 8.56 (0.90) & 8.89 (0.98) & 8.01 (0.87) & 8.80 (1.02) & 8.43 (1.00) \\
Y mean & 8.75 (0.81) & 8.95 (0.77) & 9.05 (0.93) & 8.92 (0.78) & 8.98 (0.87) \\
X range & 7.71 (1.74) & 7.73 (1.81) & 8.23 (1.78) & 8.05 (1.81) & 7.97 (1.62) \\
Y range & 10.16 (1.62) & 10.81 (1.65) & 9.68 (1.68) & 10.18 (1.72) & 9.96 (1.65) \\
Sample size & 220 & 220 & 220 & 220 & 220 \\
\hline
\end{tabular}
\begin{tablenotes}[flushleft]
\footnotesize
\item[] Note: All values in all the statistics tables are reported as Mean (Standard Deviation) except Sample Size which indicates the number of samples per class.
\end{tablenotes}
\end{threeparttable}
\end{table}

\begin{table}[htbp]
\centering
\small
\caption{Descriptive statistics for Letters F to J}
\label{tab:stemnist_stats_2}
\begin{tabular}{@{}lccccc@{}}
\hline
Character & \textbf{F} & \textbf{G} & \textbf{H} & \textbf{I} & \textbf{J} \\
\hline
ON events & 60.83 (15.00) & 68.83 (21.15) & 64.93 (18.35) & 61.69 (19.98) & 60.70 (17.94) \\
OFF events & 63.00 (16.24) & 67.61 (19.81) & 66.22 (18.80) & 63.74 (21.82) & 61.68 (17.77) \\
X mean & 8.03 (1.10) & 8.73 (0.93) & 9.07 (0.97) & 8.78 (1.05) & 8.45 (1.11) \\
Y mean & 7.67 (0.74) & 9.20 (0.74) & 8.98 (0.75) & 9.27 (0.90) & 9.32 (1.02) \\
X range & 7.27 (1.88) & 7.54 (1.71) & 6.97 (1.66) & 7.31 (2.23) & 7.51 (1.76) \\
Y range & 9.80 (1.53) & 10.07 (1.60) & 10.14 (1.60) & 9.59 (1.44) & 10.40 (1.55) \\
Sample size & 220 & 220 & 220 & 220 & 220 \\
\hline
\end{tabular}
\end{table}

\begin{table}[htbp]
\centering
\small
\caption{Descriptive statistics for Letters K to O}
\label{tab:stemnist_stats_3}
\begin{tabular}{@{}lccccc@{}}
\hline
Character & \textbf{K} & \textbf{L} & \textbf{M} & \textbf{N} & \textbf{O} \\
\hline
ON events & 65.79 (18.17) & 50.76 (15.42) & 74.66 (18.56) & 66.45 (18.95) & 66.66 (21.63) \\
OFF events & 66.82 (17.64) & 52.71 (15.76) & 75.99 (18.38) & 69.12 (21.22) & 67.09 (21.75) \\
X mean & 8.39 (0.96) & 8.07 (0.92) & 9.18 (0.93) & 8.93 (0.87) & 8.81 (0.79) \\
Y mean & 9.40 (0.87) & 10.45 (0.82) & 8.79 (1.03) & 9.02 (0.92) & 8.59 (0.94) \\
X range & 7.00 (1.70) & 7.65 (1.93) & 9.15 (1.83) & 8.14 (1.99) & 8.36 (1.74) \\
Y range & 9.93 (1.62) & 9.37 (1.45) & 9.66 (1.73) & 9.61 (1.70) & 9.63 (1.58) \\
Sample size & 220 & 220 & 220 & 220 & 220 \\
\hline
\end{tabular}
\end{table}

\begin{table}[htbp]
\centering
\small
\caption{Descriptive statistics for Letters P to T}
\label{tab:stemnist_stats_4}
\begin{tabular}{@{}lccccc@{}}
\hline
Character & \textbf{P} & \textbf{Q} & \textbf{R} & \textbf{S} & \textbf{T} \\
\hline
ON events & 63.60 (17.55) & 69.76 (18.22) & 72.33 (20.86) & 63.67 (20.20) & 55.49 (18.73) \\
OFF events & 65.44 (17.71) & 70.16 (18.19) & 72.78 (21.04) & 64.02 (19.98) & 57.82 (19.36) \\
X mean & 7.97 (1.17) & 8.82 (0.96) & 8.84 (1.17) & 8.69 (0.91) & 8.97 (0.90) \\
Y mean & 7.67 (0.84) & 9.14 (0.90) & 8.81 (0.74) & 9.14 (0.87) & 7.73 (1.04) \\
X range & 6.91 (1.65) & 8.08 (1.64) & 7.18 (1.72) & 7.20 (1.50) & 8.70 (2.14) \\
Y range & 10.28 (1.40) & 9.62 (1.52) & 10.14 (1.64) & 9.97 (1.56) & 9.87 (1.61) \\
Sample size & 220 & 220 & 220 & 220 & 220 \\
\hline
\end{tabular}
\end{table}

\begin{table}[htbp]
\centering
\small
\caption{Descriptive statistics for Letters U to Y}
\label{tab:stemnist_stats_5}
\begin{tabular}{@{}lccccc@{}}
\hline
Character & \textbf{U} & \textbf{V} & \textbf{W} & \textbf{X} & \textbf{Y} \\
\hline
ON events & 61.19 (14.99) & 57.98 (17.92) & 76.28 (20.76) & 62.93 (19.81) & 63.75 (20.42) \\
OFF events & 63.35 (15.18) & 60.65 (18.06) & 78.45 (22.20) & 64.47 (20.52) & 63.63 (19.88) \\
X mean & 8.78 (1.05) & 9.03 (0.83) & 8.93 (0.86) & 8.58 (0.89) & 8.95 (0.82) \\
Y mean & 9.20 (0.76) & 8.83 (0.81) & 9.20 (0.73) & 9.37 (0.85) & 8.91 (0.89) \\
X range & 7.36 (1.59) & 7.81 (1.93) & 9.60 (1.76) & 7.80 (1.66) & 7.48 (1.77) \\
Y range & 10.06 (1.40) & 9.90 (1.49) & 9.31 (1.74) & 9.58 (1.57) & 11.39 (1.57) \\
Sample size & 220 & 220 & 220 & 220 & 220 \\
\hline
\end{tabular}
\end{table}

\begin{table}[htbp]
\centering
\small
\caption{Descriptive statistics for Letter Z and Digits 1 to 4}
\label{tab:stemnist_stats_6}
\begin{tabular}{@{}lccccc@{}}
\hline
Character & \textbf{Z} & \textbf{1} & \textbf{2} & \textbf{3} & \textbf{4} \\
\hline
ON events & 71.97 (22.50) & 46.60 (19.41) & 63.65 (18.45) & 65.21 (18.69) & 68.74 (18.75) \\
OFF events & 75.00 (24.49) & 47.69 (19.24) & 65.52 (18.89) & 66.11 (17.92) & 71.69 (20.80) \\
X mean & 8.58 (0.92) & 8.55 (0.99) & 8.71 (0.74) & 9.03 (0.81) & 8.86 (0.97) \\
Y mean & 9.08 (0.85) & 9.47 (1.06) & 9.69 (0.99) & 9.26 (1.01) & 9.23 (0.83) \\
X range & 8.31 (1.64) & 4.02 (3.10) & 8.20 (1.56) & 7.22 (1.68) & 8.45 (1.82) \\
Y range & 8.75 (1.59) & 10.06 (1.67) & 8.95 (1.58) & 10.33 (1.65) & 10.54 (1.69) \\
Sample size & 220 & 220 & 220 & 220 & 220 \\
\hline
\end{tabular}
\end{table}

\begin{table}[htbp]
\centering
\small
\caption{Descriptive statistics for Digits 5 to 9}
\label{tab:stemnist_stats_7}
\begin{tabular}{@{}lccccc@{}}
\hline
Character & \textbf{5} & \textbf{6} & \textbf{7} & \textbf{8} & \textbf{9} \\
\hline
ON events & 65.26 (19.26) & 59.52 (17.75) & 53.53 (17.38) & 72.05 (19.49) & 69.71 (18.69) \\
OFF events & 67.21 (19.52) & 59.31 (17.23) & 53.91 (17.14) & 72.77 (19.62) & 69.42 (17.46) \\
X mean & 8.55 (0.87) & 8.40 (0.94) & 9.21 (0.87) & 8.71 (0.88) & 9.26 (1.03) \\
Y mean & 8.61 (0.78) & 9.37 (0.91) & 7.99 (1.02) & 8.75 (0.94) & 7.93 (1.09) \\
X range & 7.85 (1.69) & 6.45 (1.67) & 6.91 (1.68) & 6.23 (1.57) & 6.34 (1.68) \\
Y range & 9.81 (1.39) & 9.80 (1.41) & 10.09 (1.39) & 10.16 (1.47) & 11.15 (1.66) \\
Sample size & 220 & 220 & 220 & 220 & 220 \\
\hline
\end{tabular}
\end{table}

Tables \ref{tab:stemnist_stats_1}-\ref{tab:stemnist_stats_7} present detailed statistics for each character class, following the format established by ST-MNIST. High standard deviations in these statistics indicate that there is significant variation in spike counts across all samples within a class. The reason for this variance lies in the different ways people write, the speed of their writing and the amount of pressure they put down. Relative complexities and stroke count differences between these characters can be clearly reflected with the help of this example. Class `B' has 78.1$\pm$23.5 ON events and 78.6$\pm$22.9 OFF events on average, whereas class `L' has far fewer 50.8$\pm$15.4 ON events and 52.7$\pm$15.8 OFF events. Spatial statistics (X mean, Y mean, X range and Y range) also demonstrate a great deal of variability within and between character classes. The design of STEMNIST does not impose any spatial constraints on the starting and ending point of the recordings, allowing a great amount of variability of human handwriting to be preserved - as opposed to N-MNIST, which was derived from pre-centered static MNIST images with fixed spatial extents. The variance of writing positions, character sizes and orientations inherent in unrestricted tactile handwriting is thus captured by the large standard deviations for spatial range statistics.

\begin{figure}[htb]
\centering
\includegraphics[width=\textwidth]{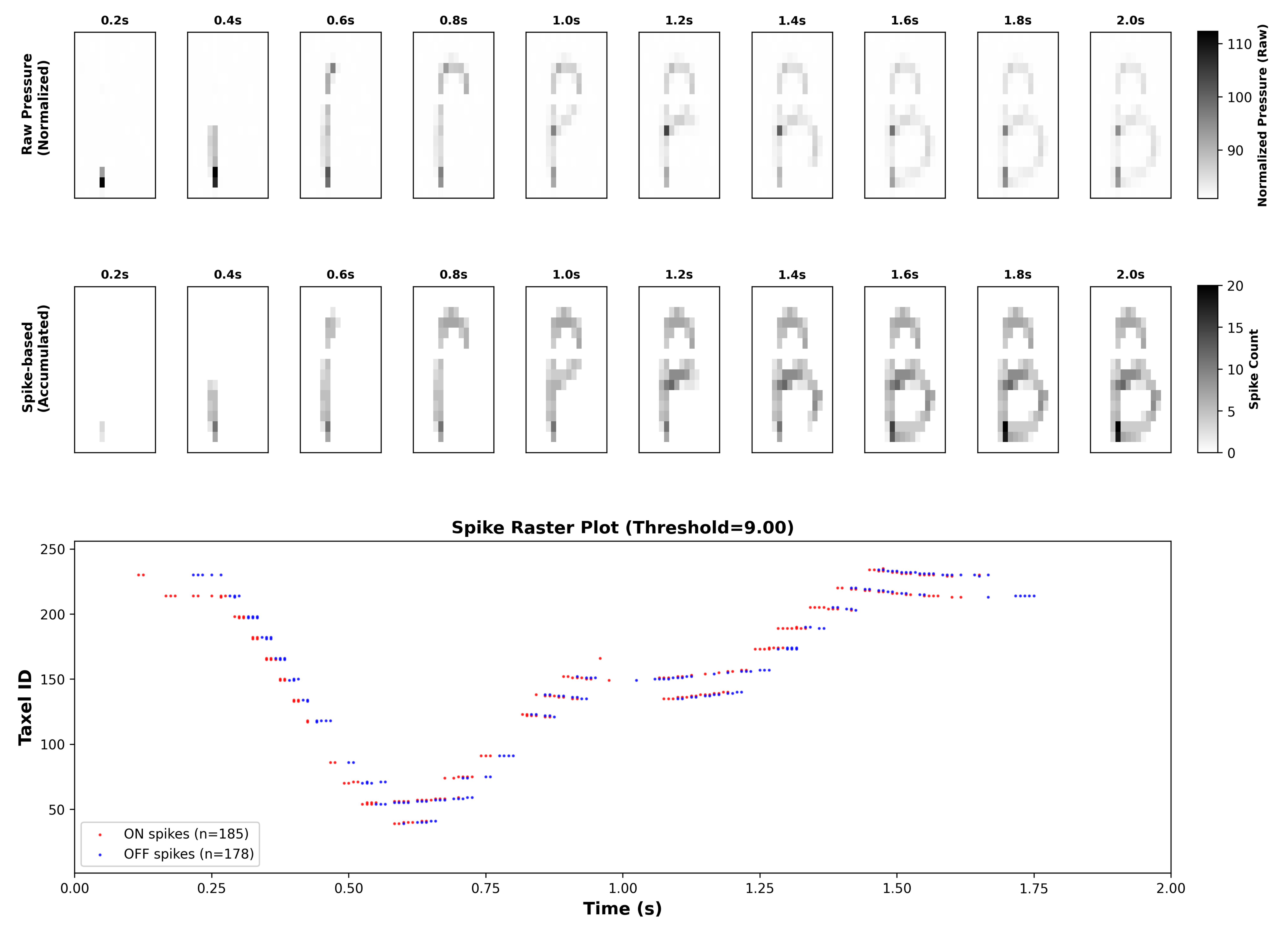}
\caption[Evolution of tactile representation for letter B]{
\textbf{Top Row:} Evolution of the normalized ``tactile image'' over time for the letter `B', derived from accumulated raw pressure frames. \textbf{Middle Row:} Temporal evolution of the spike-based tactile representation, obtained by summing count of ON and OFF spike events over time. \textbf{Bottom Row:} Spike raster plot showing 185 ON events (red) and 178 OFF events (blue), along the 256 taxels in the duration of the 2-second recording. For this sample, the adaptive threshold was $\theta_{\text{sample}} = 9.0$.}
\label{fig:evolution_spike_raster}
\end{figure}

The temporal evolution of tactile portrayal for a representative sample of the letter `B' is depicted in Figure \ref{fig:evolution_spike_raster}. The top row shows the normalized tactile images which are based on time-integrated raw pressure data and illustrate the gradual build-up of the character stroke trajectory. The middle row illustrates the images based on spike events that have accumulated over time spatially. These accession of ON and OFF events show how the event-driven representation preserves the underlying spatio-temporal structure of patterning of the handwriting gesture. The bottom spike raster plot displays the actual temporal distribution of 185 ON events and 178 OFF events from the 256 taxels over the 2 second recording window. The vertical stem, upper loop and lower loop components of the letter `B' are represented by distinct temporal clusters, while the concentration of spike activity in particular taxel ranges correlates to the sequential strokes that form the character. The event-based representation supplies rich spatiotemporal features suitable for neuromorphic spiking neural network processing by effectively capturing how the strokes related to tactile handwriting unfold over time and space.

Depending on the specific character being written, these statistics can be used as discriminative features for classification tasks. However, previous work\cite{orchard2015converting} on neuromorphic datasets indicates that if summary statistics as features for a simple classifier (e.g., k-Nearest Neighbors) are used directly, performance is observed to be sub-optimal compared to a classifier which utilizes the full spatio-temporal structure of a spike train. Accordingly, the classification experiments discussed in Section \ref{sec:classification} focus on making use of the full spike trains instead of just summary statistics.

\section{Classification}\label{sec:classification}
The core tactile recognition task studied in this work is multi-class classification over the set of 35 classes comprising uppercase letters \textbf{A-Z} and digits \textbf{1-9}. This direct analogy to the EMNIST protocol provides a substantially more challenging and representative benchmark for evaluation than the 10-class digit recognition task of ST-MNIST, requiring models to discriminate between topologically similar characters sharing overlapping stroke primitives (e.g., B/P/R/K, C/O/Q, I/J/L, U/V/W). To establish baseline performance metrics and validate the discriminability of our event-based tactile dataset, we implemented two classification approaches: a conventional Convolutional Neural Network (CNN) processing dual-channel spike accumulation maps, and a neuromorphic-compatible Spiking Neural Network (SNN) exploiting temporal spike dynamics. Our evaluation protocol consists of three stages: (1) hyperparameter optimization via 5-fold cross-validation on the training set, (2) robustness assessment through multi-split training with different random data partitions and (3) final held-out test set evaluation following strict machine learning best practices to prevent data leakage.

\subsection{Model Architectures}
\paragraph{Convolutional Neural Network (CNN):}
The CNN employs a compact feedforward design optimized for the 16$\times$16 spatial resolution of the tactile sensor array. The input consists of a 2-channel spike accumulation map, where channel 0 encodes ON events (positive pressure changes) and channel 1 encodes OFF events (negative pressure changes), aggregated across the 2-second recording window. This dual-polarity representation preserves bidirectional tactile dynamics while enabling efficient processing through standard convolution operations. The architecture comprises two convolutional blocks followed by a fully connected classifier:
\begin{itemize}
    \item \textbf{Conv Block 1:} 8 filters with 4$\times$4 kernel, stride 1, no padding (output: 8$\times$13$\times$13), Rectified Linear Unit (ReLU) activation, 2$\times$2 max pooling with stride 2 (output: 8$\times$6$\times$6).
    \item \textbf{Conv Block 2:} 16 filters with 3$\times$3 kernel, stride 1, padding 1 (output: 16$\times$6$\times$6), ReLU activation, 2$\times$2 max pooling with stride 2 (output: 16$\times$3$\times$3).
    \item \textbf{Dropout:} Regularization layer with drop probability 0.3 applied to the 144-dimensional flattened feature vector.
    \item \textbf{Classifier:} Fully connected layer mapping 144 features to 35 output classes (26 letters A-Z + 9 digits 1-9).
\end{itemize}
The total parameter count is 6,507 (Conv1: 264, Conv2: 1,168, FC: 5,075), ensuring computational efficiency suitable for edge deployment on resource-constrained systems.

\paragraph{Spiking Neural Network (SNN):}
The SNN architecture mirrors the CNN's spatial topology but replaces static ReLU activations with Leaky Integrate-and-Fire (LIF) neurons to process temporal spike sequences directly. Input spike trains are binned into 20 discrete time steps over the 2-second recording window (100 ms temporal resolution per bin), preserving the asynchronous dynamics of tactile handwriting strokes. Each convolutional and fully connected layer is followed by a LIF neuron population with membrane decay constant, $\beta = 0.9$, modeling bio-realistic spike-timing-dependent dynamics. The forward pass iterates through all 20 time steps sequentially, with membrane potentials integrating spike inputs and generating output spikes when exceeding the firing threshold. Final classification is performed by summing output spike counts across the temporal dimension using rate coding. The architecture is as follows:
\begin{itemize}
    \item \textbf{Conv Block 1:} 8 filters with 4$\times$4 kernel, LIF neurons with $\beta=0.9$, 2$\times$2 max pooling.
    \item \textbf{Conv Block 2:} 16 filters with 3$\times$3 kernel, LIF neurons with $\beta=0.9$, 2$\times$2 max pooling.
    \item \textbf{Dropout:} Regularization with drop probability 0.3 applied to pooled features before classification.
    \item \textbf{Classifier:} Fully connected layer to 35 classes with output LIF neurons; spike counts summed over time steps for classification.
\end{itemize}
Training employed snnTorch\cite{eshraghian2023training} for surrogate gradient-based backpropagation through time, using the arctangent (ATan) surrogate gradient function, which is snnTorch's default gradient approximation for the non-differentiable Heaviside spike function. The ATan surrogate provides a smooth, bell-shaped gradient that balances training stability and gradient magnitude across a wide range of membrane potentials, enabling end-to-end gradient descent despite the non-differentiable spike generation mechanism. Both architectures share identical parameter counts (6,507) to ensure fair comparison of representational capacity. All experiments were conducted on an NVIDIA GPU with CUDA acceleration using PyTorch 2.0. No data augmentation (e.g., rotation, scaling, noise injection) was applied to preserve naturalistic tactile handwriting variability.

\subsection{Hyperparameter Optimization}\label{sec:hpo}
To identify optimal training configurations while preventing test set contamination, we performed 5-fold stratified cross-validation exclusively on the training partition (6,160 samples, 80\% of the full dataset). Each fold maintained balanced class distributions, with 4,928 training samples and 1,232 validation samples per fold. Multiple hyperparameter combinations were evaluated for both architectures, exploring learning rates $\{0.001, 0.005\}$, batch sizes $\{32, 64\}$, and for SNN specifically, temporal resolutions $\{10, 20\}$ time steps. All configurations incorporated dropout regularization (probability 0.3) and L2 weight decay ($10^{-4}$) to prevent overfitting.

The CNN was trained for 20 epochs per fold, while the SNN required 50 epochs due to the increased complexity of learning temporal spike patterns through surrogate gradients. Training employed cross-entropy loss and Adam optimizer. The best hyperparameter configuration for CNN, selected by highest mean 5-fold validation accuracy, was: learning rate 0.005, batch size 64, achieving \textbf{89.72\% $\pm$ 0.61\%} cross-validation accuracy. For SNN, optimal settings were: learning rate 0.005, batch size 64, 20 time steps, yielding \textbf{88.25\% $\pm$ 0.98\%} cross-validation accuracy. The narrow 1.47\% performance gap at this stage indicated that the dual-channel spike encoding and temporal binning strategy effectively preserved discriminative tactile features for neuromorphic processing.

\subsection{Multi-Split Robustness Evaluation}
\begin{figure}[htb]
    \centering
    \includegraphics[width=\textwidth]{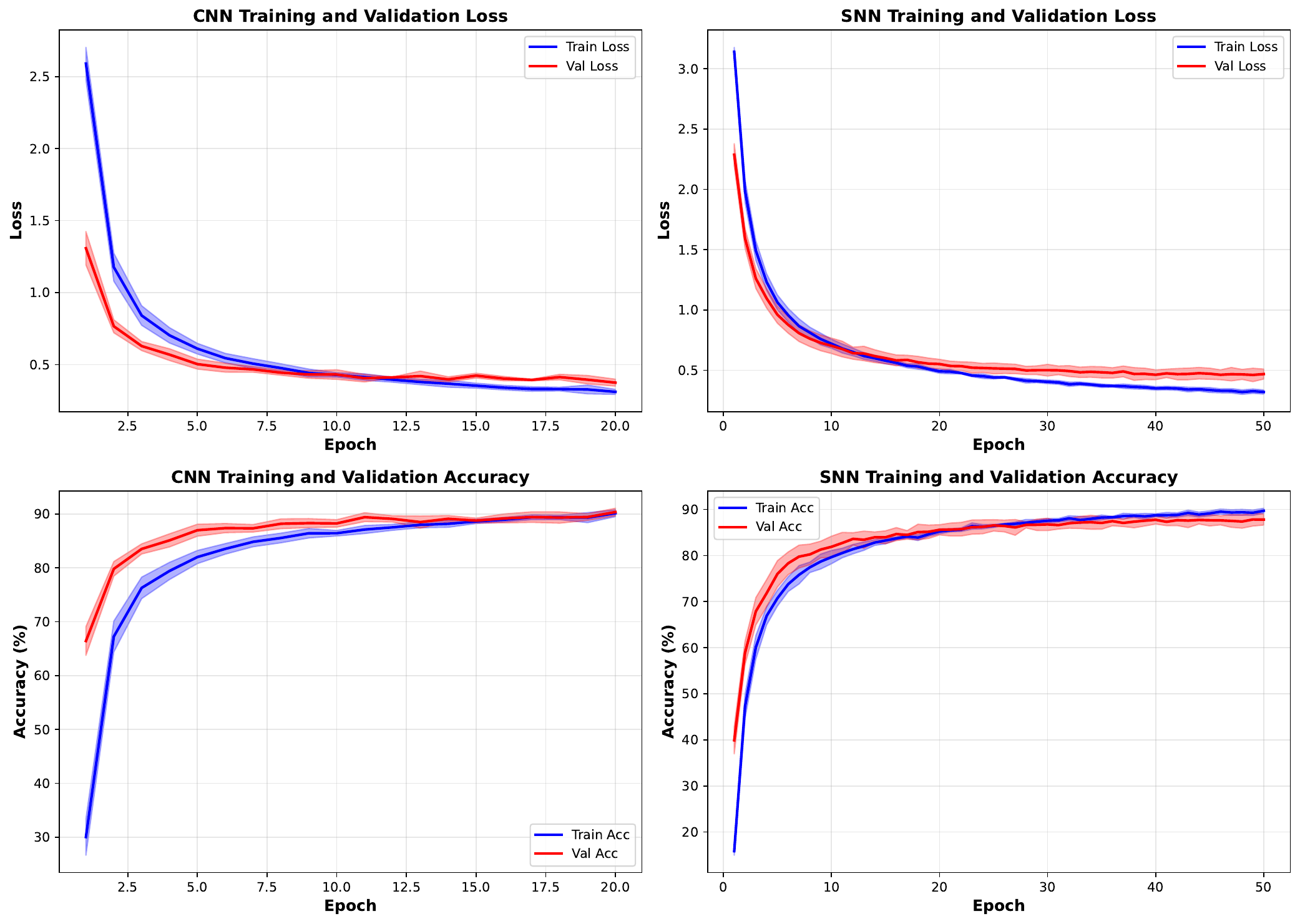}
    \caption{Learning curves averaged across five random sub-splits. \textbf{Top Row:} Training and validation loss. \textbf{Bottom Row:} Training and validation accuracy. Shaded regions indicate $\pm$1 standard deviation. CNN (left) converges in 20 epochs; SNN (right) requires 50 epochs for comparable performance. Dropout and weight decay prevent overfitting, as evidenced by stable validation loss.}
    \label{fig:learning_curves}
\end{figure}

To assess model stability and generalization consistency beyond a single train-test partition, we trained five independent models for each architecture using different random 80-20 splits of the training set, employing random seeds $\{42, 123, 456, 789, 1011\}$. Each split preserved stratified class balance, and both model weights and random number generators were independently reinitialized per split. This protocol ensures reported performance metrics reflect robustness to data sampling variability rather than fortuitous partitioning. Using the optimal hyperparameters identified in \ref{sec:hpo}, we trained each model from scratch and tracked epoch-wise training loss, validation loss, training accuracy, and validation accuracy. These metrics were aggregated across the five splits to compute mean learning curves with standard deviation bands, visualizing convergence behavior and variance (Figure \ref{fig:learning_curves}). The CNN achieved \textbf{90.36\% $\pm$ 0.70\%} mean validation accuracy across splits (range: 89.04 -  91.07\%), while the SNN reached \textbf{87.81\% $\pm$ 1.20\%} (range: 85.80 - 89.29\%). The larger variance in SNN performance reflects known challenges in surrogate gradient training sensitivity to initialization, though both architectures demonstrated stable convergence with decreasing or plateauing validation loss, confirming effective regularization.

\subsection{Final Test Set Evaluation}
Following hyperparameter selection and robustness validation, we trained the final models on the complete 6,160-sample training set (80\% of full dataset) using optimal configurations and evaluated once on the held-out 1,540-sample test set which was never accessed during any of the earlier stages. This strict protocol prevents information leakage and provides unbiased performance estimates representative of real-world deployment scenarios.

\textbf{Quantitative Performance:} The final CNN achieved \textbf{90.91\% test accuracy}, while the SNN reached \textbf{89.16\%}, yielding a minimal \textbf{1.75\%} performance gap. Both models substantially outperform random guessing (2.86\% for 35 equiprobable classes) and even exceed the accuracies reported by See et al.\cite{see2020st} for 10-class ST-MNIST. Notably, both of our 35-class CNN and SNN accuracies surpass the 89\% ANN performance on 10-class ST-MNIST despite presenting a 3.5$\times$ harder classification problem, demonstrating the effectiveness of systematic regularization and dual-channel encoding.

\begin{figure}[htb]
    \centering
    \includegraphics[width=\textwidth]{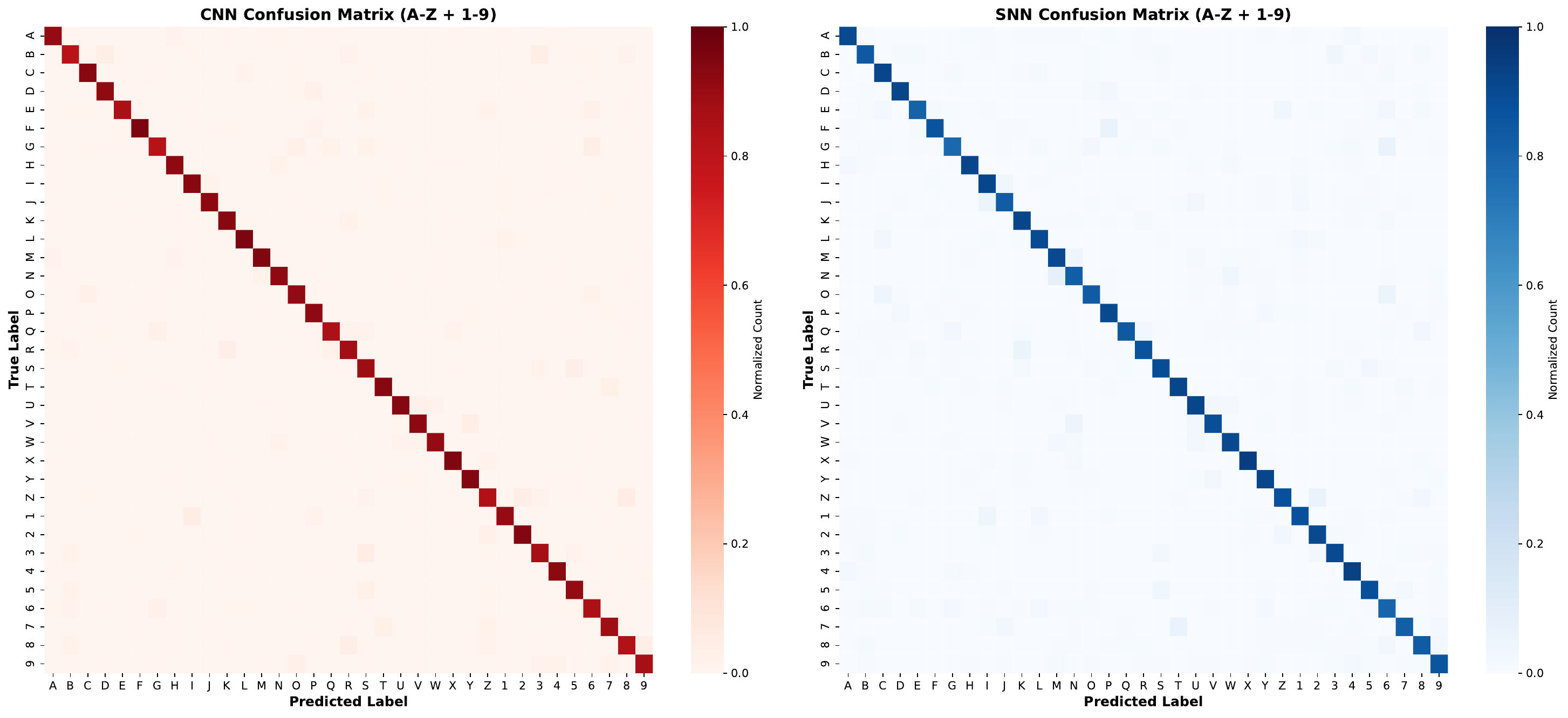}
    \caption{Normalized confusion matrices averaged across five sub-split models. \textbf{Left:} CNN. \textbf{Right:} SNN. Both exhibit strong diagonal dominance, with common errors between visually similar characters (e.g., I/J/L for letters, 5/6/8 for digits), consistent with human perceptual ambiguities \cite{vega1991human} observed in tactile letter recognition tasks, where topologically similar characters sharing stroke primitives (vertical stems, curved segments) produce asymmetric confusion patterns even among trained participants.}
    \label{fig:confusion_matrices}
\end{figure}

\textbf{Confusion Analysis:} Figure \ref{fig:confusion_matrices} presents normalized confusion matrices averaged across the five sub-split validation sets, revealing class-wise performance patterns. Both architectures exhibit strong diagonal dominance, indicating high per-class accuracy, with misclassifications concentrated among topologically similar character pairs: I/J/L (vertical strokes), M/N/W (multi-peak structures), U/V (curved strokes), and digits 5/6/8 (shared loop components). These error patterns align with expected human perceptual confusions in unrestricted tactile handwriting and highlight remaining challenges for real-world deployment.

\begin{figure}[htb]
    \centering
    \includegraphics[width=\textwidth]{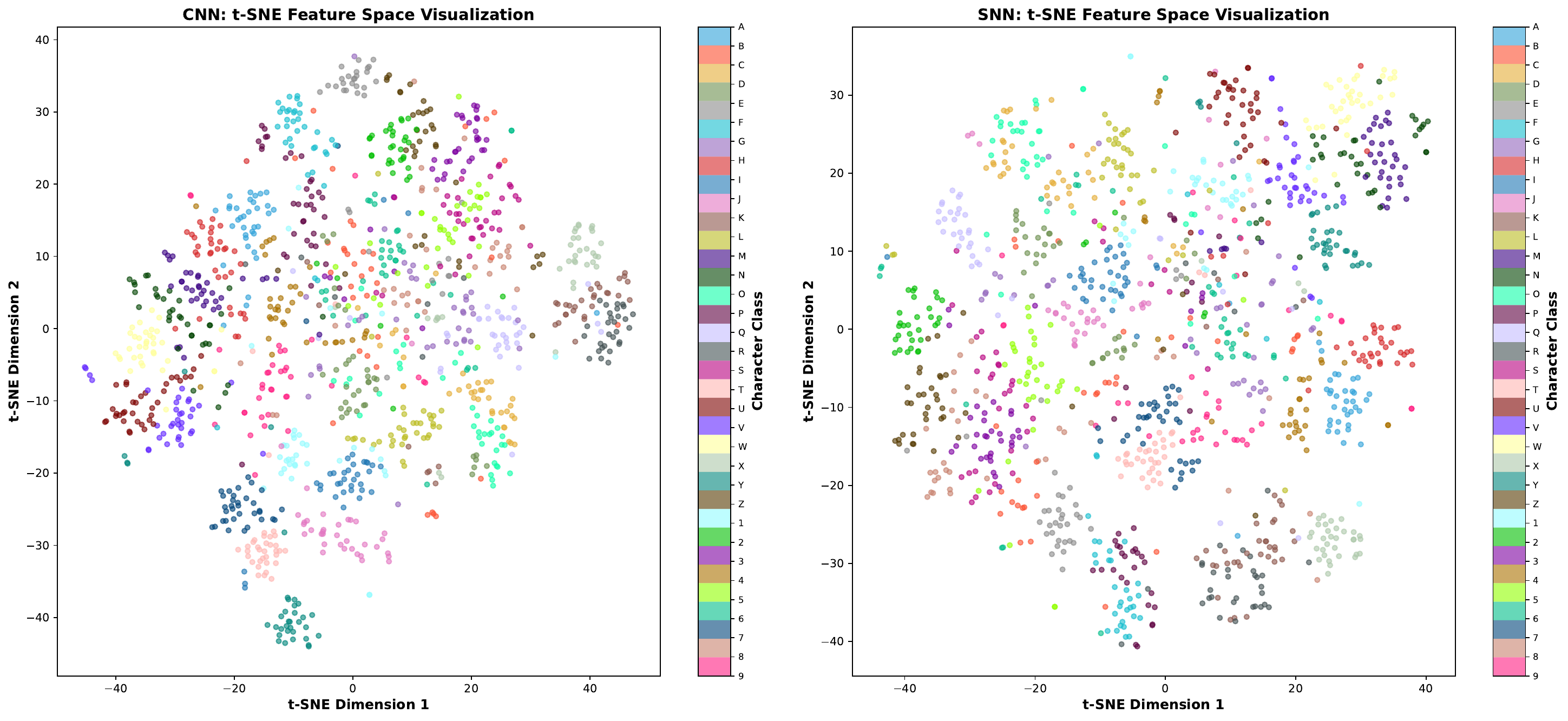}
    \caption{Feature Space Visualization using t-SNE of penultimate layer features (144 dimensions) extracted from one validation split. \textbf{Left:} CNN features. \textbf{Right:} SNN features.}
    \label{fig:tsne}
\end{figure}

\textbf{Feature Space Visualization:} To qualitatively assess learned representations, we applied t-distributed Stochastic Neighbor Embedding (t-SNE\cite{maaten2008visualizing}) to 144-dimensional penultimate-layer features extracted from one validation split (Figure \ref{fig:tsne}).For the CNN, features were obtained directly from the flattened post-pooling activations (16$\times$3$\times$3 = 144 dimensions). For the SNN, we extracted features by summing membrane potentials across all 20 timesteps for each neuron in the penultimate layer, following the rate-coding paradigm where temporal spike counts (proportional to integrated membrane potentials) encode class information. This temporal integration collapses the 20$\times$144 spatiotemporal feature tensor into a single 144-dimensional vector per sample, enabling direct comparison with CNN features in the same embedding space. Both CNN and SNN produce well-separated clusters for most character classes, with some expected overlap between similar characters (e.g., U/V/W, K/R/Y). The SNN feature space shows slightly greater cluster diffusion compared to CNN, likely reflecting the stochastic nature of spike-timing dynamics and temporal integration variability across the 2-second recording window. Nevertheless, clear class boundaries persist, confirming that temporally-integrated spike patterns encode sufficient discriminative information for high-accuracy neuromorphic classification.

\section{Future Directions}\label{sec:future}
These baselines establish STEMNIST as a challenging neuromorphic benchmark for tactile recognition and point to few performance gaps compared to ST-MNIST, thus offering ample opportunities for algorithmic innovation and dataset expansion. Future directions lie across three complementary domains: real-world applications using event-based tactile sensing; dataset extensions to broaden coverage and modality; and algorithmic developments that exploit spatio-temporal spike dynamics.

\subsection{Potential Applications}
The event-based nature of this dataset positions it for deployment in several neuromorphic computing scenarios. This dataset can help assistive technologies for visually impaired users, such as tactile Braille readers or haptic text interfaces, bridge the gap between traditional vision-based Optical Character Recognition (OCR) and touch-based recognition with a 35-class alphanumeric character set. Human Machine Interfaces and prosthetics can utilize neuromorphic tactile feedback in order to provide closed-loop sensory restoration with improved latency and minimal power consumption. Training energy-efficient event recognition pipelines on neuromorphic hardware using the dataset could also provide real-time haptic feedback for dexterous tasks in unstructured environments to robotic tactile perception systems.

\subsection{Dataset Extensions}
This work can be extended in future iterations by increasing class diversity in an effort to improve generalization. The class count can be increased to 61 by including lowercase letters from `a' to `z', and the addition of zero (0) can give it full parity with EMNIST. This enables mixed-case handwriting recognition. Adding multilingual characters, such as Devanagari script, Chinese strokes, or Bengali numerals, extends the applications to global contexts. Synchronizing these tactile recordings with vision through camera capture of tool trajectory allows for multi-modal data fusion for cross-modal learning and sensor fusion benchmarks. 

\subsection{Algorithmic Developments}
The modest performance difference between CNN and SNN motivates to work towards several algorithmic innovations. Deeper architectures with residual connections, batch normalization and adaptive LIF thresholds can better exploit spatial features and temporal dynamics. Recurrent Neural Networks such as LSTM-SNN hybrids can model long-range temporal dependencies across stroke sequences, capturing stroke order and inter-stroke timing. Attention mechanisms for spike trains such as temporal attention over a suitable number of time bins or spatial attention over the 16$\times$16 array can dynamically weight discriminative features. Self-supervised pretraining on unlabeled tactile data such as contrastive learning on stroke embeddings can improve feature representations before fine-tuning. Graph neural networks may capture structural priors by modeling stroke topology, considering nodes as taxels and edges as temporal connectivity. Neuromorphic hardware deployment with on-chip learning rules, such as spike timing-dependent plasticity, eligibility traces may enable real-time energy-efficient training. Hybrid ANN-SNN conduits, combining the spatial feature extraction from CNN with SNN temporal processing, can exploit their complementary strengths. Lastly, few-shot and meta-learning can be used to approach participant-specific handwriting variations with minimal per-user calibration data.

\section{Conclusion}\label{sec:conclusion}
We introduce STEMNIST, the first large-scale neuromorphic tactile dataset, expanding the original ST-MNIST framework from 10 digits to 35 alphanumeric classes (A-Z and 1-9), addressing a critical gap in event-based haptic benchmarks for neuromorphic computing. The dataset consists of 7,700 samples collected from 34 participants using a custom-built 16$\times$16 tactile sensor array operating at 120 Hz, yielding 1,005,592 spike events encoded through adaptive thresholding. This extension presents significantly higher classification difficulty compared to the earlier digit-only datasets and offers the complexity of EMNIST's visual character recognition challenge within the tactile domain. Baseline classification experiments using conventional CNNs and spiking neural networks demonstrate both the discriminative power of the dataset and the effectiveness of dual-channel spike encoding for bio-inspired tactile processing. Through systematic hyperparameter optimization via 5-fold cross-validation (achieving 89.72\% $\pm$ 0.61\% for CNN and 88.25\% $\pm$ 0.98\% for SNN), followed by robustness evaluation across five independent data splits (90.36\% $\pm$ 0.70\% CNN, 87.81\% $\pm$ 1.20\% SNN), our final models achieved 90.91\% and 89.16\% test accuracy respectively maintaining a minimal 1.75\% CNN-SNN performance gap, compared to typical 8\%+ gaps reported in prior neuromorphic character recognition studies. Both Neural Networks achieve test accuracies significantly above random chance and even surpass the CNN and SNN baselines for the ST-MNIST classification performance. STEMNIST closes the gap between reduced tactile digit recognition and more complex real-world haptic interaction scenarios that require discrimination of alphanumeric characters. The event-based format, unrestricted spatial variability and rich temporal dynamics make the dataset an ideal testbed for the deployment of neuromorphic hardware as well as bio-inspired learning algorithms, including STDP, recurrent SNNs and attention mechanisms. The standardized train-test splits, enriched statistics and baseline benchmarks here enable the reproducibility of evaluations for tactile recognition studies on conventional deep learning, neuromorphic computing and hybrid architectures. In addition, this dataset will aim to grow in scope and applicability with future extensions to lowercase letters, multilingual characters and multi-modal fusions. We hope that STEMNIST will help catalyze the development of energy-efficient tactile perception in robotics, assistive technologies and human-machine interfaces toward the broader vision of Neuromorphic Artificial Intelligence with efficiency and adaptability rivaling those of biological sensory processing.

\section{Dataset Availability}
The STEMNIST Dataset along with supporting documentation can be accessed here:

\section{Acknowledgments}
The authors express their gratitude to all those who contributed to the collection of the STEMNIST dataset as well as the discussions at the Bangalore Neuromorphic Engineering Workshop, 2025 where the project was conceived. This work was supported by the Research Grants Council of the HK SAR, China (Project No. CityU 11212823).

\section*{References}
\bibliography{refs.bib}

\begin{thebibliography}{10}
\providecommand{\url}[1]{#1}
\csname url@samestyle\endcsname
\providecommand{\newblock}{\relax}
\providecommand{\bibinfo}[2]{#2}
\providecommand{\BIBentrySTDinterwordspacing}{\spaceskip=0pt\relax}
\providecommand{\BIBentryALTinterwordstretchfactor}{4}
\providecommand{\BIBentryALTinterwordspacing}{\spaceskip=\fontdimen2\font plus
\BIBentryALTinterwordstretchfactor\fontdimen3\font minus \fontdimen4\font\relax}
\providecommand{\BIBforeignlanguage}[2]{{%
\expandafter\ifx\csname l@#1\endcsname\relax
\typeout{** WARNING: IEEEtran.bst: No hyphenation pattern has been}%
\typeout{** loaded for the language `#1'. Using the pattern for}%
\typeout{** the default language instead.}%
\else
\language=\csname l@#1\endcsname
\fi
#2}}
\providecommand{\BIBdecl}{\relax}
\BIBdecl

\bibitem{Lederman2009HapticTutorial}
\BIBentryALTinterwordspacing
S.~J. Lederman and R.~L. Klatzky, ``Haptic perception: A tutorial,'' \emph{Attention, Perception, \& Psychophysics}, vol.~71, no.~7, pp. 1439--1459, 2009. [Online]. Available: \url{https://pubmed.ncbi.nlm.nih.gov/19801605/}
\BIBentrySTDinterwordspacing

\bibitem{gibson1962observations}
J.~J. Gibson, ``Observations on active touch.'' \emph{Psychological review}, vol.~69, no.~6, p. 477, 1962.

\bibitem{jin2023progress}
J.~Jin, S.~Wang, Z.~Zhang, D.~Mei, and Y.~Wang, ``Progress on flexible tactile sensors in robotic applications on objects properties recognition, manipulation and human-machine interactions,'' \emph{Soft Science}, vol.~3, no.~1, pp. N--A, 2023.

\bibitem{roberts2021soft}
P.~Roberts, M.~Zadan, and C.~Majidi, ``Soft tactile sensing skins for robotics,'' \emph{Current Robotics Reports}, vol.~2, no.~3, pp. 343--354, 2021.

\bibitem{wangwei_kilo}
W.~W. Lee, S.~Kukreja, and N.~Thakor, ``A kilohertz kilotaxel tactile sensor array for investigating spatiotemporal features in neuromorphic touch,'' in \emph{2015 IEEE Biomedical Circuits and Systems Conference (BioCAS)}.\hskip 1em plus 0.5em minus 0.4em\relax IEEE, 2015, pp. 1--4.

\bibitem{tactile_temporal_neurosci}
J.~A. Pruszynski and R.~S. Johansson, ``Edge-orientation processing in first-order tactile neurons,'' \emph{Nature Neuroscience}, vol.~17, p. 1404–1409, 2014.

\bibitem{lecun2002gradient}
Y.~LeCun, L.~Bottou, Y.~Bengio, and P.~Haffner, ``Gradient-based learning applied to document recognition,'' \emph{Proceedings of the IEEE}, vol.~86, no.~11, pp. 2278--2324, 2002.

\bibitem{cohen2017emnist}
G.~Cohen, S.~Afshar, J.~Tapson, and A.~Van~Schaik, ``Emnist: Extending mnist to handwritten letters,'' in \emph{2017 international joint conference on neural networks (IJCNN)}.\hskip 1em plus 0.5em minus 0.4em\relax IEEE, 2017, pp. 2921--2926.

\bibitem{see2020st}
H.~H. See, B.~Lim, S.~Li, H.~Yao, W.~Cheng, H.~Soh, and B.~C. Tee, ``St-mnist--the spiking tactile mnist neuromorphic dataset,'' \emph{arXiv preprint arXiv:2005.04319}, 2020.

\bibitem{sundaram2019learning}
S.~Sundaram, P.~Kellnhofer, Y.~Li, J.-Y. Zhu, A.~Torralba, and W.~Matusik, ``Learning the signatures of the human grasp using a scalable tactile glove,'' \emph{Nature}, vol. 569, no. 7758, pp. 698--702, 2019.

\bibitem{muller2022braille}
S.~F. M{\"u}ller-Cleve, V.~Fra, L.~Khacef, A.~Peque{\~n}o-Zurro, D.~Klepatsch, E.~Forno, D.~G. Ivanovich, S.~Rastogi, G.~Urgese, F.~Zenke \emph{et~al.}, ``Braille letter reading: A benchmark for spatio-temporal pattern recognition on neuromorphic hardware,'' \emph{Frontiers in Neuroscience}, vol.~16, p. 951164, 2022.

\bibitem{grother1995nist}
P.~J. Grother, ``Nist special database 19,'' \emph{Handprinted forms and characters database, National Institute of Standards and Technology}, vol.~10, p.~69, 1995.

\bibitem{serrano2015poker}
T.~Serrano-Gotarredona and B.~Linares-Barranco, ``Poker-dvs and mnist-dvs. their history, how they were made, and other details,'' \emph{Frontiers in neuroscience}, vol.~9, p. 481, 2015.

\bibitem{orchard2015converting}
G.~Orchard, A.~Jayawant, G.~K. Cohen, and N.~Thakor, ``Converting static image datasets to spiking neuromorphic datasets using saccades,'' \emph{Frontiers in neuroscience}, vol.~9, p. 437, 2015.

\bibitem{posch2008asynchronous}
C.~Posch, D.~Matolin, and R.~Wohlgenannt, ``An asynchronous time-based image sensor,'' in \emph{2008 IEEE International Symposium on Circuits and Systems (ISCAS)}.\hskip 1em plus 0.5em minus 0.4em\relax IEEE, 2008, pp. 2130--2133.

\bibitem{iyer2021neuromorphic}
L.~R. Iyer, Y.~Chua, and H.~Li, ``Is neuromorphic mnist neuromorphic? analyzing the discriminative power of neuromorphic datasets in the time domain,'' \emph{Frontiers in neuroscience}, vol.~15, p. 608567, 2021.

\bibitem{amir2017low}
A.~Amir, B.~Taba, D.~Berg, T.~Melano, J.~McKinstry, C.~Di~Nolfo, T.~Nayak, A.~Andreopoulos, G.~Garreau, M.~Mendoza \emph{et~al.}, ``A low power, fully event-based gesture recognition system,'' in \emph{Proceedings of the IEEE conference on computer vision and pattern recognition}, 2017, pp. 7243--7252.

\bibitem{kong2025super}
D.~Kong, Y.~Lu, S.~Zhou, M.~Wang, G.~Pang, B.~Wang, L.~Chen, X.~Huang, H.~Lyu, K.~Xu \emph{et~al.}, ``Super-resolution tactile sensor arrays with sparse units enabled by deep learning,'' \emph{Science Advances}, vol.~11, no.~27, p. eadv2124, 2025.

\bibitem{rueckauer2017conversion}
B.~Rueckauer, I.-A. Lungu, Y.~Hu, M.~Pfeiffer, and S.-C. Liu, ``Conversion of continuous-valued deep networks to efficient event-driven networks for image classification,'' \emph{Frontiers in neuroscience}, vol.~11, p. 682, 2017.

\bibitem{eshraghian2023training}
J.~K. Eshraghian, M.~Ward, E.~O. Neftci, X.~Wang, G.~Lenz, G.~Dwivedi, M.~Bennamoun, D.~S. Jeong, and W.~D. Lu, ``Training spiking neural networks using lessons from deep learning,'' \emph{Proceedings of the IEEE}, vol. 111, no.~9, pp. 1016--1054, 2023.

\bibitem{vega1991human}
F.~Vega-Bermudez, K.~O. Johnson, and S.~S. Hsiao, ``Human tactile pattern recognition: active versus passive touch, velocity effects, and patterns of confusion,'' \emph{Journal of neurophysiology}, vol.~65, no.~3, pp. 531--546, 1991.

\bibitem{maaten2008visualizing}
L.~v.~d. Maaten and G.~Hinton, ``Visualizing data using t-sne,'' \emph{Journal of machine learning research}, vol.~9, no. Nov, pp. 2579--2605, 2008.

\end{thebibliography}
\bibliographystyle{IEEEtran}
\csname url@samestyle\endcsname
\providecommand{\newblock}{\relax}
\providecommand{\bibinfo}[2]{#2}
\providecommand{\BIBentrySTDinterwordspacing}{\spaceskip=0pt\relax}
\providecommand{\BIBentryALTinterwordstretchfactor}{4}
\providecommand{\BIBentryALTinterwordspacing}{\spaceskip=\fontdimen2\font plus
\BIBentryALTinterwordstretchfactor\fontdimen3\font minus
  \fontdimen4\font\relax}
\providecommand{\BIBforeignlanguage}[2]{{%
\expandafter\ifx\csname l@#1\endcsname\relax
\typeout{** WARNING: IEEEtran.bst: No hyphenation pattern has been}%
\typeout{** loaded for the language `#1'. Using the pattern for}%
\typeout{** the default language instead.}%
\else
\language=\csname l@#1\endcsname
\fi
#2}}
\providecommand{\BIBdecl}{\relax}
\BIBdecl

\end{document}